\documentclass[review]{elsarticle}

\usepackage{hyperref}

\usepackage{amsmath}
\usepackage{graphicx}
\usepackage{multirow}
\usepackage[lofdepth,lotdepth]{subfig}
\usepackage{float}
\usepackage{rotating}

\begin{document}

\begin{frontmatter}

\title{Incremental Minimax Optimization based Fuzzy Clustering for Large Multi-view Data}


\author[mymainaddress]{Yangtao Wang}

\author[mysecondaryaddress]{Lihui Chen}

\author[mymainaddress]{Xiaoli Li}

\address[mymainaddress]{Data Analytics Department, Institute for Infocomm Research, A*STAR, 1 Fusionopolis Way, 138632, Singapore}
\address[mysecondaryaddress]{School of Electrical and Electronic Engineering, Nanyang Technological University, 50 Nanyang Ave, 639798, Singapore\corref{mycorrespondingauthor}}

\begin{abstract}
Incremental clustering approaches have been proposed for handling large data when given data set is too large to be stored. The key idea of these approaches is to find representatives to represent each cluster in each data chunk and final data analysis is carried out based on those identified representatives from all the chunks. However, most of the incremental approaches are used for single view data. As large multi-view data generated from multiple sources becomes prevalent nowadays, there is a need for incremental clustering approaches to handle both large and multi-view data. In this paper we propose a new incremental clustering approach called incremental minimax optimization based fuzzy clustering (IminimaxFCM) to handle large multi-view data.

 In IminimaxFCM, representatives with multiple views are identified to represent each cluster by integrating multiple complementary views using minimax optimization. The detailed problem formulation, updating rules derivation, and the in-depth analysis of the proposed IminimaxFCM are provided. Experimental studies on several real world multi-view data sets have been conducted. We observed that IminimaxFCM outperforms related incremental fuzzy clustering in terms of clustering accuracy, demonstrating the great potential of IminimaxFCM for large multi-view data analysis.
\end{abstract}


\end{frontmatter}


\section{Introduction}
Clustering as an important data analysis technique to find information underlining the unlabelled data, is widely applied in different areas including data mining, machine learning and information retrieval. Different clustering algorithms based on various theories have been developed over the past decades \cite{jain2010data,filippone2008survey,xu2005survey}. To handle large data which is too large to be stored, incremental clustering approaches have been proposed. In incremental clustering, data is processed chunk by chunk, which is a packet of the data. The key idea of these approaches is to find representatives to represent each cluster in each data chunk and final data analysis is carried out based on those identified representatives from all the chunks. Most of the incremental clustering approaches are applied to handle single view data in which each object is represented by single kind of features. However, large multi-view data becomes prevalent nowadays because huge amount of data can be collected easily every day from different sources or represented by different features. For example, one image can be represented by different kinds of features including color, texture and histogram .etc. which can be considered as different views. As each view can provide distinctive yet complementary information, mining large multi-view data is important for different parties including enterprises and government organizations. Under this circumstance, there is a need to develop incremental clustering approaches to handle both large and multi-view data. For large multi-view data clustering, the key challenge is that how to categorize the data set in the incremental clustering framework by making good use of related information from multiple views of the data. In this paper we propose a new incremental clustering approach called incremental minimax optimization based fuzzy clustering (IminimaxFCM) to handle large multi-view data. In IminimaxFCM, the entire data set is processed chunk by chunk. As every data object in each chunk has multiple views, minimax optimization is applied to integrate multiple views to achieve consensus results.

For large data clustering, different strategies have been applied in the literature. For example random sampling strategy is used in CLARA \cite{Kaufman2009} and CLARANS \cite{Ng:1994:EEC:645920.672827} to handle large data. Summarization strategy is applied in BIRCH \cite{livny1996birch}. In \cite{guha2003clustering}, the incremental clustering strategy is used for stream data based on k-Median called LSEARCH is developed using chunk-based processing style. In \cite{chitta2011approximate,chitta2012efficient} random sampling and random Fourier maps based kernel k-means are proposed to perform large scale kernel clustering. Except the hard clustering discussed above, similar strategies are also used in soft or fuzzy clustering approaches in order to handle more real world data sets where data objects may not be well separated. It has been discussed that soft clustering such as those popular ones in the literature \cite{mei2012fuzzy} may capture the natural structure of a data set more accurately. Each object in a data set may belong to all clusters with various degrees of memberships. For example, incremental clustering strategy is also applied in an online Nonnegative Matrix Factorization approach \cite{wang2011efficient} for large scale document clustering. In addition, several incremental fuzzy clustering algorithms based on the well known Fuzzy c means (FCM) \cite{Bezdek:1981:PRF:539444} and Fuzzy c medoids (FCMD) \cite{krishnapuram2001low} have been developed respectively. The popular algorithms include the single-pass FCM (SPFCM) \cite{hore2007single}, online FCM (OFCM) \cite{hore2008online,hore2009scalable}, online fuzzy c medoids (OFCMD) \cite{labroche2010new} and history based online fuzzy c medoids (HOFCMD) \cite{labroche2010new}. To improve the clustering performance, incremental fuzzy clustering with multiple medoids \cite{wang2014incremental} is proposed. However, both hard and soft clustering approaches above belong to single view clustering approaches in which the data is represented by single view features or single relational matrix, hence they are not suitable to handle large multi-view data.

To handle large multi-view data, multi-view clustering technique can be integrated into the incremental clustering framework. There are many multi-view clustering approaches have been proposed in the literature. Three strategies are mainly applied among the existing approaches. The first strategy is to formulate an objective function by minimizing the disagreements of different views and optimize it directly to get the consensus clustering results \cite{2011NIPScosc, 2012ICDMmvKernelKmeans, 2013IJCAIRmultiviewKmeans}. The second strategy applied in \cite{2012cvpraffinity, 2013AAAIconvexSubspaceMultiview} contains two steps as follows. First, a unified representation (view) is identified or learned. Then the existing clustering algorithm such as K-means \cite{macqueen1967some} or spectral clustering \cite{ng2002spectral} is used to get the final clustering result. In the third strategy \cite{bruno2009multiview, greene2009matrix}, each view of the data is processed independently and the consensus clustering result is achieved in an additional step based on the result of each view.

The above approaches are also based on hard clustering in which each object is assigned to only one cluster. However in many real world applications more information about the cluster assignment need to be captured. For example, in document categorization each document may belong to several topics with different degrees of attachments. Moreover, each document may be written by different languages or collected from various sources. Therefore, several soft or fuzzy clustering based multi-view clustering approaches are proposed. For example, Nonnegative Matrix Factorization (NMF) \cite{lee1999learning} based approach has been proposed in \cite{liu2013multi}. Fuzzy clustering based approaches are also proposed including CoFKM \cite{2009COFKM} and WV-Co-FCM \cite{2014WVCOFCM}. In \cite{wang2014multi}, minimax optimization based multi-view spectral clustering is proposed to handle multi-view relational data. The experimental results have shown that minimax optimization helps to integrate different views more effectively. However, all these approaches assume that the data set can be stored and processed in batch, hence they are not suitable for clustering large multi-view data which may be too large to be stored in the memory.

To handle large multi-view vector data, we propose a new incremental clustering approach called incremental minimax optimization based fuzzy clustering (IminimaxFCM). To the best of our knowledge, this is the first research effort made in developing incremental fuzzy clustering approach for large multi-view data based on minimax optimization. Moreover, inspired by recent advancements in minimax optimization and fuzzy clustering, IminimaxFCM can integrate different views better and also produce soft assignment for each object. In IminimaxFCM, the data is processed one chunk at a time instead of loading the entire data into the memory for analysis. Multi-view centroids for each cluster in a data chunk are identified based on minimax optimization by minimizing the maximum disagreement of the weighted views. These multi-view centroids are then used as the representatives of the multi-view data to carry out the final data partition.   The detailed formulation, derivation and an in-depth analysis of the approach are given in the paper. The experiments of IminimaxFCM on six real world data sets show that IminimaxFCM achieves better clustering accuracy than related incremental fuzzy clustering approaches.

The rest of the paper is organized as follows: in the next section, a review on the related incremental fuzzy approaches reported in the literature is highlighted. In section 3, the details of the proposed approach IminimaxFCM are presented. Comprehensive experiments on several real world data sets are conducted and the detailed comparison results with related techniques are reported in section 4.  Finally, conclusions are drawn in section 5.

\section{Related work}
In this section, two related single view based incremental fuzzy clustering algorithms are reviewed. Their common and unique characteristics are discussed as well.
\subsection{SPFCM}
SPFCM \cite{hore2007single} is designed based on FCM \cite{Bezdek:1981:PRF:539444}.
The objective function of FCM $J_{FCM}$ is defined as follows:
 \begin{equation}
 J_{FCM} = \sum\limits_{c=1}^{k}\sum\limits_{i=1}^{n}u_{ci}^m\|x_{i}-v_{c}\|^2
 \end{equation}
 Where, $m > 1$ is the fuzzifier, $k$ is the number of clusters, $n$ is the number of objects. $u_{ci}$ is the membership of object $i$ in cluster $c$ and $v_{c}$ is the centroid of c-th cluster.
As FCM is not able to handle data when it is too large to be stored in memory, SPFCM is proposed to process the data chunk by chunk.
For each chunk,  a set of centroids is calculated to represent the chunk with one centroid per cluster. In SPFCM, the centroids identified from the previous chunk are combined into next chunk and the final set of centroids for the entire data is generated after last chunk is processed. Instead of applying FCM directly on each chunk, weighted FCM(wFCM) is used for SPFCM to determine the final set of centroids.

The significant difference of SPFCM and OFCM is the way of handling the centroids of each chunk.
In SPFCM,
the weight for the centroid of each cluster in each chunk is calculated as follows:
 \begin{equation}
  w_{c} = \sum\limits_{i=1}^{n_{p}+m}(u_{ci})w_{i},\quad
 1\leq c \leq k
 \end{equation}
 Where, $w_{c}$ is the weight of centroid of the c-th cluster,  $n_{p}$ is the number of objects in $p_{th}$ chunk, $k$ is the number of clusters, $u_{ci}$ is the membership of object $i$ belongs to cluster $c$ and $w_{i}$ is the weight of object $i$. For the first chunk of the data ($p=1$), $w_{i}$ is assigned to 1 for every object and $m = 0$. From the second chunk of data ($p\neq1$), $m = k$ and the $k$ weighted centroids are combined with the p-th chunk of data. The $n_{p}+k$ objects will be clustered by wFCM in which the weights of the $n_{p}$ objects in p-th chunk are all set to 1 and the weights of $k$ objects are calculated from previous chunk.
 These steps continue until last chunk of data is processed.
\subsection{OFCM}
Similar with SPFCM, OFCM \cite{hore2008online} is also developed based on FCM and the data is processed chunk by chunk. However, there are two significant differences between OFCM and SPFCM. The first difference is the way of handling the centroids of each chunk. In OFCM, every chunk is processed independently and an additional step is needed to identify the final set centroids for the entire data based on all the centroids identified from the chunks. The second difference is the way to calculate the weight of each cluster in a chunk.
 In OFCM, the weight is calculated as follows:
 \begin{equation}
 w_{c} = \sum\limits_{i=1}^{n_{p}}(u_{ci})w_{i},\quad
 1\leq c \leq k,\quad w_{i}=1, \quad\forall 1\leq i \leq {n_{p}}
 \end{equation}
 The weighted centroids are then input into wFCM to identify the final set of centroids for the entire data set.

As discussed above, both of the methods are applied for single view data which is represented by single view features, however they are not able to handle large multi-view data. In our method, we integrate multi-view clustering technique based on minimax optimization into incremental clustering framework to identify multi-view centroids for each chunk separately. We then identify the final set of centroids based on all the multi-view centroids from all the chunks. Next, we propose our novel incremental minimax optimization based fuzzy clustering approach called IminimaxFCM, including the detailed formulation, derivation and an in-depth analysis.

\section{The proposed approach}
In this section, we first propose two naive and simple incremental multi-view clustering approaches based on OFCM and SPFCM. We refer to these two approaches as NaiveMVOFCM and NaiveMVSPFCM. Next, we formulate the objective function of the proposed approach IminimaxFCM. In IminimaxFCM, the consensus clustering results and the multi-view centroids of each chunk are generated based on minimax optimization in which the maximum disagreements of different weighted views are minimized. Moreover, the weight of each view can be learned automatically in the clustering process. It is shown that the IminimaxFCM clustering is actually formulated as a minimax optimization problem of the cost function with constraints.  The Lagrangian Multiplier method is applied to derive the updating rules. Finally we introduce the algorithm of IminimaxFCM including the detailed step. The time complexity of the algorithm will be discussed as well. In summary, multi-view centroids are identified by IminimaxFCM to represent each cluster in a chunk. The final set of multi-view centroids can be identified at last.
\subsection{NaiveMVOFCM and NaiveMVSPFCM}
We first propose NaiveMVOFCM and NaiveMVSPFCM in which OFCM and SPFCM are applied in each view of the data set to achieve a set of centroids to represent each view. Then the final label of each data object is determined based on the its distance to the centroids of each view. The details of the approach are shown in Algorithm 1 and Algorithm 2 respectively. Though these two methods are able to handle large multi-view data, the clustering performance may not good because they have not used the information among complementary views to achieve a consensus clustering result. Next, we propose the method called IMinimaxFCM by integrating different views based on minimax optimization.
\begin{center}
\begin{tabular}{ll}
\hline
\textbf{Algorithm 1:} NaiveMVOFCM \\\hline
\textbf{Input:} Data set of $P$ views $\{X^{(1)}, ... X^{(P)}\}$ with size $N$ \\
\quad\quad\quad Cluster Number $K$, stopping criterion $\epsilon$, fuzzifier $m$\\

\textbf{Output:} Cluster Indicator $\textbf{q}$\\
\textbf{Method:} \\
\quad {\scriptsize 1} \textbf{for} $p=1 \,\text{to} \,P$ \\
\quad\quad\quad    Identify centroids $v_{c}^{p}$ for each view based on $X^{(P)}$ \\
\quad\quad\quad    by using OFCM \\
\quad\,\, \textbf{end for}\\
\quad{\scriptsize 2} \textbf{for} $j=1 \,\text{to} \,N$\\
\quad\quad\quad    $\textbf{q}_{j} = arg\,min_{c \in \{1,2,...K\}} \sum\limits_{p=1}^{P}\parallel x_{j}^{p}-v_{c}^{p}\parallel, $\,\,\\
\quad\,\, \textbf{end for}\\
\hline
\end{tabular}
\end{center}

\begin{center}
\begin{tabular}{ll}
\hline
\textbf{Algorithm 2:} NaiveMVSPFCM \\\hline
\textbf{Input:} Data set of $P$ views $\{X^{(1)}, ... X^{(P)}\}$ with size $N$ \\
\quad\quad\quad Cluster Number $K$, stopping criterion $\epsilon$, fuzzifier $m$\\

\textbf{Output:} Cluster Indicator $\textbf{q}$\\
\textbf{Method:} \\
\quad {\scriptsize 1} \textbf{for} $p=1 \,\text{to} \,P$ \\
\quad\quad\quad    Identify centroids $v_{c}^{p}$ for each view based on $X^{(P)}$ \\
\quad\quad\quad    by using SPFCM \\
\quad\,\, \textbf{end for}\\
\quad{\scriptsize 2} \textbf{for} $j=1 \,\text{to} \,N$\\
\quad\quad\quad    $\textbf{q}_{j} = arg\,min_{c \in \{1,2,...K\}} \sum\limits_{p=1}^{P}\parallel x_{j}^{p}-v_{c}^{p}\parallel, $\,\,\\
\quad\,\, \textbf{end for}\\
\hline
\end{tabular}
\end{center}
\subsection{IminimaxFCM}
The objective function based on minimax optimization is formulated for IminimaxFCM $J_{IminimaxFCM}$ as follows:
 \begin{equation}
 J_{IminimaxFCM} = \min\limits_{U^{\ast}, \{V^{(p)}\}_{p=1}^{P}}\quad\max\limits_{ \{\alpha^{(p)}\}_{p=1}^{P}}\;\sum\limits_{p=1}^{P}(\alpha^{(p)})^{\gamma}Q^{(p)}
\label{equation:JMVFC}
\end{equation}
here,
\begin{equation}
Q^{(p)} = \sum\limits_{c=1}^{K}\sum\limits_{i=1}^{N}(u_{ci}^{\ast})^{m}\|x_{i}^{(p)}-v_{c}^{(p)}\|^2
\label{equation:QP}
\end{equation}
subject to
\begin{equation}
\sum\limits_{c=1}^{K}u_{ci}^{\ast}= 1, \text{\,for\,}  i = 1,2,...,N
\label{equation:uci}
\end{equation}
\begin{align}
u_{ci}^{\ast}\geq 0, \text{\,for\,} c = 1,2,...,K,  i = 1,2,...N
\label{equation:uci2}
\end{align}
\begin{equation}
\sum\limits_{p=1}^{P}\alpha^{(p)}= 1
\label{equation:alphap}
\end{equation}
\begin{align}
\alpha^{(p)}\geq 0, \text{\,for\,} p = 1,2,...,P
\label{equation:alphap2}
\end{align}
 In the objective function,
 $U^{\ast}$ is the $K\times N$ membership matrix contains the consensus membership $u_{ci}^{\ast}$. $V^{(p)}$ is the $D^{(p)}\times K$ centroid matrix of $p_{th}$ view contains the centroids to represent the data of $p_{th}$ view. Here $D^{(p)}$ is the dimension of the objects in $p_{th}$ view.
 $Q^{(p)}$ can be considered as the cost of  $p_{th}$ view which is the standard objective function of FCM. $(\alpha^{(p)})^{\gamma}$ is the weight of $p_{th}$ view.  The parameter $\gamma \in [0, 1)$ controls the distribution of weights $(\alpha^{(p)})^{\gamma}$ for different views. $m>1$ is the fuzzifier for fuzzy clustering which controls the fuzziness of the membership.

 The goal is to conduct a minimax optimization on the objective function $\sum\limits_{p=1}^{P}(\alpha^{(p)})^{\gamma}Q^{(p)}$ with the constraints to get the consensus membership and the multi-view centroids to represent the data. In this new formulation for multi-view clustering, the weights for each view are automatically determined based on minimax optimization, without specifying the weights by users. Moreover, by using  minimax optimization, the different views are integrated harmonically by weighting each cost term $Q^{(p)}$ differently.
 \subsection{Optimization}
 It is difficult to solve the variables $u_{ci}^{\ast}$, $v_{c}^{(p)}$ and $\alpha^{(p)}$ in (\ref{equation:JMVFC}) directly because (\ref{equation:JMVFC}) is nonconvex. However, as we observed that the objective function is convex w.r.t $u_{ci}^{\ast}$ and $v_{c}^{(p)}$ and is concave w.r.t  $\alpha^{(p)}$, therefore, similar to FCM, the alternative optimizaiton(AO) can be used to solve the optimization problem by solving one variable with others fixed.
  \subsubsection{Minimization: Fixing $v_{c}^{(p)}$, $\alpha^{(p)}$and updating $u_{ci}^{\ast}$}
  Lagrangian Multiplier method is applied to solve the optimization problem of the objective function with constraints. The Lagrangian function considering the constraints is given as follows:
 \begin{equation}
\begin{split}
L_{IminimaxFCM}
 &=J_{IminimaxFCM} \\
 &+ \sum\limits_{i=1}^{N}\lambda_{i}(\sum\limits_{c=1}^{K}u_{ci}^{\ast}-1)
  + \beta(\sum\limits_{p=1}^{P}\alpha^{(p)}-1) \\
\end{split}
\label{equation:LSSPFC}
\end{equation}
where the $J_{IminimaxFCM}$ is the objective function of IminimaxFCM $\sum\limits_{p=1}^{P}(\alpha^{(p)})^{\gamma}Q^{(p)}$. $\lambda_{i}$ and $\beta$ are the Lagrange multipliers. The condition for solving $u_{ci}^{\ast}$ is as follows:
\begin{equation}
\frac{\partial_{L_{IminimaxFCM}}}{\partial_{u_{ci}^{\ast}}} = 0
\label{equation:UCIAST}
\end{equation}
Based on (\ref{equation:UCIAST}) and constraint  (\ref{equation:uci}),  the updating rule of $u_{ci}^{\ast}$ can be derived as follows:
\begin{equation}
u_{ci}^{\ast} = \left[\sum\limits_{j=1}^{K}\left(\frac{\sum\limits_{p=1}^{P}(\alpha^{(p)})^{\gamma}\|x_{i}^{(p)}-v_{c}^{(p)}\|^{2}}{\sum\limits_{p=1}^{P}(\alpha^{(p)})^{\gamma}\|x_{i}^{(p)}-v_{j}^{(p)}\|^{2}}\right)^{\frac{1}{m-1}}\right]^{-1}
\label{equation:UCIAST1}
\end{equation}
As shown in (\ref{equation:UCIAST1}), the weight $(\alpha^{(p)})^{\gamma}$ for each view is considered in the updating for $u_{ci}^{\ast}$.
\subsubsection{Minimization: Fixing $u_{ci}^{\ast}$, $\alpha^{(p)}$and updating $v_{c}^{(p)}$}
By taking derivative of $L_{IminimaxFCM}$ with respect to $v_{c}^{(p)}$, we get:
\begin{equation}
\frac{\partial_{L_{IminimaxFCM}}}{\partial_{v_{c}^{(p)}}} = -2\sum\limits_{i=1}^{N}(\alpha^{(p)})^{\gamma}(u_{ci}^{\ast})^{m}(x_{i}^{(p)}-v_{c}^{(p)})
\label{equation:VCP}
\end{equation}
The updating rule of $v_{c}^{(p)}$ is derived as follow by setting  (\ref{equation:VCP}) to be 0:
\begin{equation}
v_{c}^{(p)} = \frac{\sum\limits_{i=1}^{N}(u_{ci}^{\ast})^{m}x_{i}^{(p)}}{\sum\limits_{i=1}^{N}(u_{ci}^{\ast})^{m}}
\label{equation:VCP1}
\end{equation}
As shown in (\ref{equation:VCP1}), the updating of the centroids of each view is same as the standard FCM, however it is updated based on the consensus membership $u_{ci}^{\ast}$.
\subsubsection{Maximization: Fixing $u_{ci}^{\ast}$, $v_{c}^{(p)}$and updating $\alpha^{(p)}$}
Based on the Lagrangian Multiplier method, the condition for solving $\alpha^{(p)}$ is as follows:
\begin{equation}
\frac{\partial_{L_{IminimaxFCM}}}{\partial_{\alpha^{(p)}}} = 0
\label{equation:ALPHAP}
\end{equation}
Based on (\ref{equation:ALPHAP}) and constraint  (\ref{equation:alphap}), the updating rule $\alpha^{(p)}$ is given as:
\begin{equation}
\alpha^{(p)} = \left[\sum\limits_{j=1}^{P}\left(\frac{Q^{(p)}}{Q^{(j)}}\right)^{\frac{1}{\gamma-1}}\right]^{-1}
\label{equation:ALPHAP1}
\end{equation}
Here the cost term $Q^{(p)}$ is the weighted distance summation of all the data points under $p_{th}$ view to its corresponding centroid. The larger the value of $Q^{(p)}$ is, the larger cost this view will contribute to the objective function. From (\ref{equation:ALPHAP1}), we can see that the larger cost of $p_{th}$ view is, the higher value will be assigned to $\alpha^{(p)}$ which leads to the maximum of the weighted cost. The maximum is minimized with respect to the membership and centroids in order to suppress the high cost views and achieve a harmonic consensus clustering result.  Next, we present the details of IminimaxFCM algorithm.
\subsection{IminimaxFCM Algorithm}
\begin{figure}
\centering
\includegraphics[width=4in]{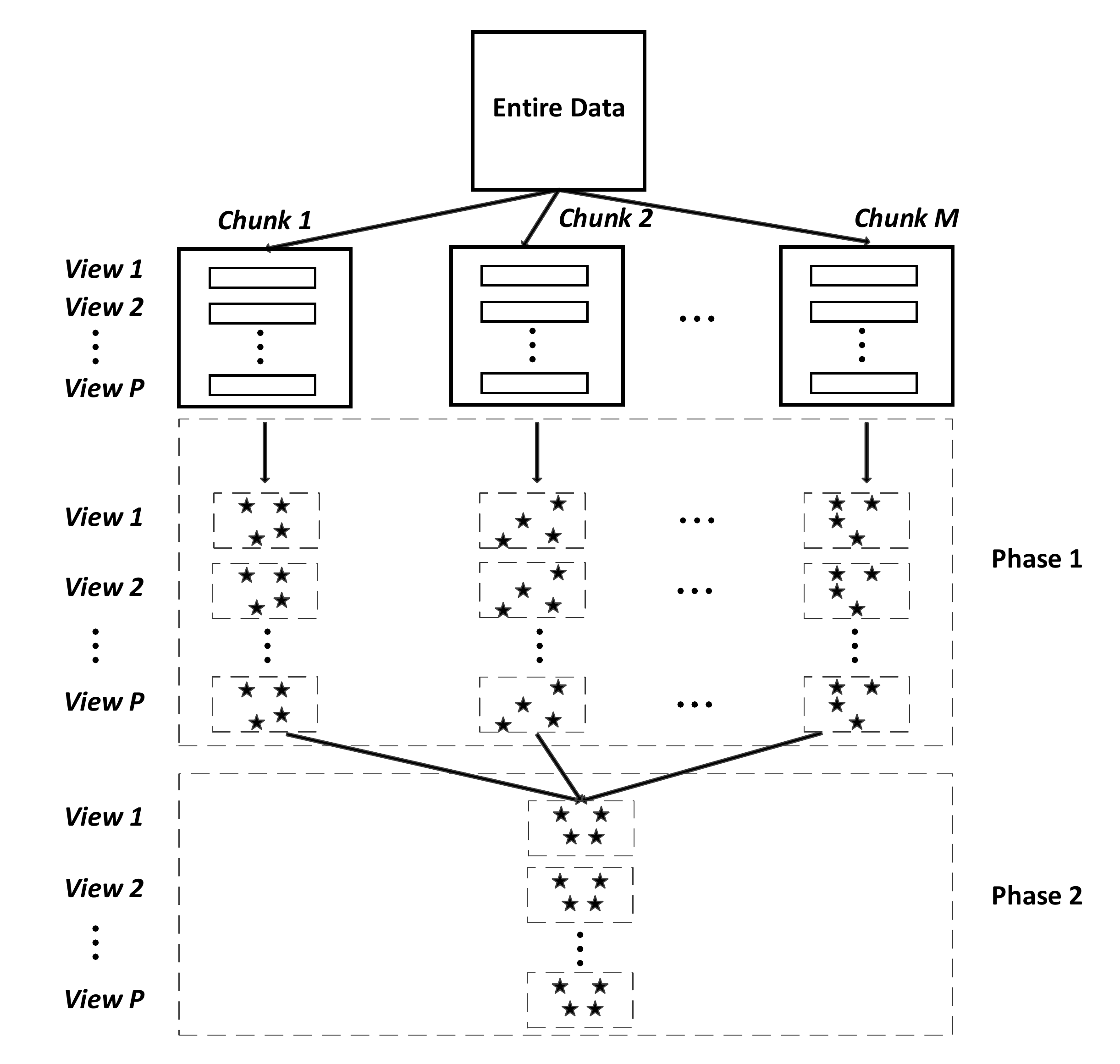}
\caption{The framework of IMinimaxFCM}
\label{fig:imvfc}
\end{figure}

The proposed IMinimaxFCM processes the large multi-view data chunk by chunk and the multi-view centroids for each chunk are identified based on minimax optimization. The framework of IMinimaxFCM is shown in Fig.~\ref{fig:imvfc}.

As shown in the framework, the entire multi-view data is divided into chunks in which each data object has $P$ views. Then in Phase 1, the centroids with $P$ views are identified independently to represent each chunk based on Algorithm 2. In Phase 2, the final set of centroids are identified to represent the entire data set by using all the centroids identified from all the chunks. The details of algorithm are outlined as follows.

First, data set $X$ is partitioned into $M$ non-overlapping chunks of $X = \{{X_{1}, X_{2},...,X_{M}}\}$ in which each object has $P$ views. In step 1, centroids set $Q_{}^{p}$ of each view are initialized. In step 2, the centroid matrix of each view $V_{m}^{(p)}$ for each chunk $m$ is calculated based on Algorithm 3 by integrating multiple views. Each column in centroid matrix $V_{m}^{(p)}$ is one centroid for a cluster in $m_{th}$ under $p_{th}$ view. In Algorithm 3, the data in each chunk with multiple views is processed. The multi-view centroids are identified by iteratively updating (\ref{equation:VCP1}), (\ref{equation:UCIAST1}) and (\ref{equation:ALPHAP}) until convergence. After each of the chunk is processed, the multi-view centroids are combined into centroids set $Q$. In step 3, the final set of centroids is identified based on $Q$ by using Algorithm 3. In the final step, cluster indicator $\textbf{q}$ is determined for each object. $\textbf{q}_{j}$ is the cluster number which object $j$ belongs to. This is acquired by assigning object $j$ to the cluster whose centroids in all $P$ views has the smallest sum of distance to object $j$.
\begin{center}
\begin{tabular}{ll}
\hline
\textbf{Algorithm:} IminimaxFCM\\\hline
\textbf{Input:} Data set of $P$ views $\{X^{(1)}, ... X^{(P)}\}$ with size $N$ \\
\quad\quad\quad Cluster Number $K$, Chunk number $M$, stopping criterion $\epsilon$, fuzzifier $m$\\

\textbf{Output:} Cluster Indicator $\textbf{q}$\\
\textbf{Method:} \\
\quad {\scriptsize 1} Initialize Centroids set for each view $Q_{}^{p}={\O}$, \\
\quad {\scriptsize 2} \textbf{for} $m=1 \,\text{to} \,M$ \\
\quad\quad\quad    Calculate centroid matrix $V_{m}^{(p)}$ for $m_{th}$ chunk of each view \\
\quad\quad\quad    based on Algorithm 3\\
\quad\quad \, \textbf{for} $p=1 \,\text{to} \,P$\\
\quad\quad\quad\quad    $Q_{}^{(p)}=Q_{}^{(p)}\cup\{V_{m}^{(p)}\}$ \\
\quad\quad\,\, \textbf{end for}\\
\quad\,\, \textbf{end for}\\
\quad{\scriptsize 3} Calculate final centroid matrix $V^{(p)}$ for each view based on Algorithm 3\\
\quad\,\,           based on all the identified centroids of $P$ views $\{Q^{(1)}, ... Q^{(P)}\}$  \\
\quad\,\,           from all the chunks \\
\quad{\scriptsize 4} \textbf{for} $j=1 \,\text{to} \,N$\\
\quad\quad\quad    $\textbf{q}_{j} = arg\,min_{c \in \{1,2,...K\}} \sum\limits_{p=1}^{P}\parallel x_{j}^{p}-v_{c}^{p}\parallel, $\,\,\\
\quad\,\, \textbf{end for}\\
\hline
\end{tabular}
\end{center}

\begin{center}
\begin{tabular}{ll}
\textbf{Algorithm:} Algorithm 3\\\hline
\textbf{Input:} Data set of $P$ views $\{X^{(1)}, ... X^{(P)}\}$ with size $N$, Cluster Number $K$ \\
\quad\quad\quad stopping criterion $\epsilon$, fuzzifier $m$, Parameter $\gamma$\\
\textbf{Output:}  multi-view centroids $V$\\
\textbf{Method:} \\
\quad {\scriptsize 1} Initialize consensus membership $u_{ci}^{\ast}$ and $\alpha^{(p)}=\frac{1}{P}$ for each view\\
\quad {\scriptsize 2} Set $t=0$\\
\quad\,\,    \textbf{Repeat} \\
\quad {\scriptsize 3}\quad     \textbf{for} $c=1 \,\text{to} \,K$ \\
\quad {\scriptsize 4}\quad\,\,    \textbf{for} $p=1 \,\text{to} \,P$ \\
\quad {\scriptsize 5}\quad\,\,    Update $v_{c}^{(p)}$ using equation (\ref{equation:VCP1}); \\
\quad {\scriptsize 6}\quad\,\,     \textbf{end for}\\
\quad {\scriptsize 7}\quad     \textbf{end for}\\
\quad {\scriptsize 8}\quad     \textbf{for} $c=1 \,\text{to} \,K$ \\
\quad {\scriptsize 9}\quad\,\,    \textbf{for} $i=1 \,\text{to} \,N$ \\
\quad {\scriptsize 10}\quad\quad    Update $u_{ci}^{\ast}$ using equation (\ref{equation:UCIAST1});  \\
\quad {\scriptsize 11}\quad\,\,     \textbf{end for}\\
\quad {\scriptsize 12}\quad     \textbf{end for}\\
\quad {\scriptsize 13}\quad     \textbf{for} $p=1 \,\text{to} \,P$ \\
\quad {\scriptsize 14}\quad\quad  Update $\alpha^{(p)}$ using equation (\ref{equation:ALPHAP}); \\
\quad {\scriptsize 15}\quad     \textbf{end for}\\
\quad {\scriptsize 16}\quad    Update $t=t+1$\\
\quad\,\,    \textbf{Until} ($\parallel (U^{\ast})^{t+1} - (U^{\ast})^{t} \parallel < \epsilon$)\\
\hline
\end{tabular}
\end{center}
The time complexity of IMinimaxFCM is $O(MP(n_{m}DK^{2})+ (|Q|PDK^{2} + KPD))$ considering the chunk number M, the dimension of object D, the view number P and cluster number K. $O(MP(n_{m}DK^{2})$ is the time complexity of processing M chunks of P views where $n_{m}$ is the number of objects in chunk m. $O(|Q|PDK^{2} + KPD)$ is the time complexity of the final clustering and the label assignment where $|Q|$ is the number of identified centroids from all the chunks.

\section{Experimental results}
In this section, experimental studies of the proposed approach are conducted on six real world data sets, including image and document data sets. The summary of the characteristics of the data sets is shown in Table.~\ref{table:summarydata} including the number of objects, the number of classes, and different features with their dimensions. More details about the data sets are introduced in the following subsections. In the experiments, we compare IMinimaxFCM with two incremental single view based approaches SPFCM, OFCM to show that if using multiple views can improve the clustering performance. We also compare with two simple incremental multi-view based approaches NaiveMVOFCM and NaiveMVSPFCM to show that if minimax optimization based IMinimaxFCM performs better.
The experiments implemented in Matlab were conducted on a PC with four cores of Intel I5-2400 with 8 gigabytes of memory.

\begin{table}[!htbp]
\scriptsize
\caption{Summary of the six data sets} 
\centering 
\begin{tabular}{|c |c |c |c |c |c|}
\hline
 View No. & MF & IS  & Caltech7/20 & Reuters & Forest \\ [0.5ex]
\hline 
1 & $Pix(240)$ & $Shape(9)$ & $Gabor(48)$ & $English(21531)$ & $Quantitative(10)$\\
2 & $Fou(76)$ & $RGB(10)$ & $WM(40)$ & $France(24892)$ & $Wilderness(4)$\\
3 & $Fac(216)$ & $-$ & $CENTRIST(254)$ & $German(34251)$ & $SoilType(40)$\\
4 & $ZER(47)$ & $-$ & $HOG(1984)$ & $Italian(15506)$ & $-$\\
5 & $KAR(64)$ & $-$ & $GIST(512)$ & $Spanish(11547)$ & $-$\\
6 & $MOR(6)$ & $-$ & $LBP(928)$ & $-$ & $-$\\
\hline 
objects \# & $2000$ & $2310$ & $1474/2386$ & $1500$ & $581012$\\
\hline
classes \# & $10$ & $7$ & $7/20$ & $6$ & $7$\\
\hline
\end{tabular}
\label{table:summarydata} 
\end{table}

\subsection{Data sets}
We compare the performance of the five algorithms on the following data sets.

Multiple features(MF)\footnote{This data set can be downloaded on https://archive.ics.uci.edu/ml/datasets/Multiple+Features.}: This data set consists of 2000 handwritten digit images (0-9) extracted from a collection of Dutch utility maps. It has 10 classes and each class has 200
images. Each object is described by 6 different views (Fourier coefficients, profile correlations, Karhunen-Love coefficients, pixel averages, Zernike moments, morphological features).

Image segmentation(IS) data set\footnote{This data set can be downloaded on https://archive.ics.uci.edu/ml/datasets/Image+Segmentation.}: This data set is composed of 2310 outdoor images which has 7 classes. Each image is represented by 19 features. The features can be considered as two views which are shape view and RGB view. The shape view consists of 9 features which describe the shape information of each image. The RGB view consists of 10 features which describe the RGB values of each image.

Caltech7/20: These two data sets are two subsets of Caltech 101 image data set \cite{fei2004learning}. Following the setting in \cite{dueck2007non}, 7 widely used classes (1474 images) including Face, Motorbikes, Dolla-Bill, Garfield, Snoopy, Stop-Sign and
Windsor-Chair are selected which is named Caltech7. A larger subset referred as Caltech20 (2386 images) is selected by choosing 20 classes including Face, Leopards, Motorbikes, Binocular, Brain, Camera, Car-Side, Dolla-Bill, Ferry, Garfield, Hedgehog, Pagoda,
Rhino, Snoopy, Stapler, Stop-Sign, Water-Lilly, Windsor-Chair, Wrench and Yin-yang. Five kinds of features are extracted from all the images including 48 dimension Gabor feature, 40 dimension wavelet moments (WM), 254 dimension CENTRIST feature, 1984 dimension HOG feature, 512 dimension GIST feature, and 928 dimension LBP feature.

Reuters: This data set contains documents originally written in five different languages (English, French, German, Spanish and Italian) and their translations \cite{amini2009learning}. This multilingual data set covers a common set of six classes. We use documents originally in English as the first view and their four translations as the other four views. We randomly sample 1500 documents from this collection with
each of the 6 classes having 250 documents.

Forest\footnote{This data set can be downloaded on https://archive.ics.uci.edu/ml/datasets/Covertype.}: This data set is from United States Geological Survey and United State Forest Service and it has 7 classes, 581012 objects. Each object is represented as a 54 dimensional feature vector. To construct multiple views, we divide 54 features to three views with each represent one kind of feature. The first view composes of the first 10 features which are quantitative features including the distances and degrees. The second view composes of the next 4 features which represent the wilderness area each tree belongs to. The third view composes of the last 40 features which represent the soil type of place where each tree grows.
\subsection{Evaluation criterion}
Three popular external criterions \emph{Accuracy} \cite{mei2012fuzzy}, \emph{F-measure} \cite{larsen1999fast}, and \emph{Normalized Mutual Information}(NMI) \cite{strehl2003cluster} are used to evaluate the clustering results, which measure the agreement of the clustering results produced by an algorithm and the ground truth. If we refer \textit{class} as the ground truth, and \textit{cluster} as the results of a clustering algorithm, the NMI is calculated as follows:
 \begin{equation}
 NMI = \frac{\sum\limits_{c=1}^{k}\sum\limits_{p=1}^{m}n_{c}^{p}log(\frac{n\cdot n_{c}^{p}}{n_{c}\cdot n_{p}})}{\sqrt{(\sum\limits_{c=1}^{k}n_{c}log(\frac{n_{c}}{n}))(\sum\limits_{p=1}^{m}n_{p}log(\frac{n_{p}}{n}))}} \\
 \label{equation:NMIcalc}
 \end{equation}
 where $n$ is the total number of objects, $n_{c}$ and $n_{p}$ are the numbers of objects in the $c_{th}$ \textit{cluster} and the $p_{th}$ \textit{class}, respectively, and $n_{c}^{p}$ is the number of common objects in \textit{class} $p$ and \textit{cluster} $c$.  For F-measure, the calculation based on precision and recall is as follows:
  \begin{equation}
 F-measure = \frac{2\cdot precision\cdot recall}{precision + recall} \\
 \label{equation:F-meaure}
 \end{equation}
 where,
  \begin{equation}
   precision = \frac{n_{c}^{p}}{n_{c}}\\
 \label{equation:precision}
 \end{equation}
 \begin{equation}
   recall = \frac{n_{c}^{p}}{n_{p}}\\
 \label{equation:recall}
 \end{equation}
 Accuracy is calculated as follows after obtaining a one-to-one match between \textit{clusters} and
\textit{classes}:
  \begin{equation}
 Accuracy = \sum\limits_{c=1}^{k}\frac{n_{c}^{j}}{n} \\
 \label{equation:Accuracy}
 \end{equation}
 where $n_{c}^{j}$ is the number of common objects in the $c_{th}$ cluster
and its matched class $j$. The higher the values of the three criterions are, the better the clustering result is. The value is equal to 1 only when the clustering result is same as the ground truth.
\subsection{Experimental setting}
For Reuters data set, same as the experimental setting in \cite{2011NIPScosc}, Probabilistic Latent Semantic Analysis(PLSA) \cite{hofmann1999probabilistic} is applied to project the data to a 100-dimensional space and the clustering approaches are conducted on the low dimensional data. For OFCM and SPFCM, as they are single view based methods, the features of all views are concatenated as the input to OFCM and SPFCM. For initialization,  we initialize the centroids by using the method adopted in \cite{krishnapuram2001low}. The object which has the minimum distance to all the other objects is selected as the first centroid. The remaining centroids are chosen consecutively by selecting the objects that maximize their minimal distance with existing centroids. This helps the centroids distribute evenly in the data space to avoid converging to a bad local optimum. For NaiveMVOFCM and NaiveMVSPFCM, OFCM and SPFCM with the same initialization method are applied in each view of the data set to achieve a set of centroids to represent each view. For IminimaxFCM, we use two methods to initialize the consensus membership $U^{\ast}$. In the first method, for each view, we first choose the centroids with the same method mentioned above. Then, if the centroid for cluster $c$ of view $p$ is object $i$, we set $u_{ci}^{p}=1$, and set the membership of the other objects to the same cluster as $0$. At last, the consensus membership $u_{ci}^{\ast}$ is calculated as the average value of all the views. The detail steps of initialization are as follows. The second initialization method applies FCM to each view and generates the membership matrix $U^{p}$ for each view. Then same as the first method, the consensus matrix $U^{\ast}$ is calculated as $U^{\ast}$ = $\frac{1}{P}\sum\limits_{p=1}^{P}U^{p}$. We refer IminimaxFCM as IminimaxFCM1 and IminimaxFCM2 for the two initialization methods accordingly.
\begin{center}
\begin{tabular}{ll}
\hline
\textbf{Initialization for IminimaxFCM} \\\hline
Set the number of clusters $k$,
consensus membership matrix is $U_{k\times{n}}^{\ast}$\\
\textbf{for} $p=1 \,\text{to} \,P$ \\
    \qquad Calculate the first centroid for $p_{th}$ view:\\
    \qquad\qquad $w=arg\,min_{1\leq{j}\leq{n}}\sum\limits_{i=1}^{n}Dis(x_{j}^{p},x_{i}^{p}); \text{\,first centroid:\,}v_{1}^{p}=x_{w}^{p}$\\
    \qquad Centroids set $V^{p}=\{{v_{1}^{p}}\}, m=1$; \\
    \qquad Fuzzy membership $u_{mw}^{p}=1, \text{\,and\,}u_{mj}^{p}=0, \text{\,with\,}j=1,2...n,j\neq{w}$ \\
    \qquad \textbf{Repeat} \\
    \qquad\qquad $m=m+1$\\
    \qquad\qquad $w=arg\,max_{1\leq{i}\leq{n};x_{i}^{p}\not\in{V^{p}}}min_{1\leq{k}\leq{|V^{p}|}}Dis(v_{k}^{p},x_{i}^{p});$\\
    \qquad\qquad $\text{\,centroid:\,}v_{m}^{p}=x_{w}^{p}$ \\
    \qquad\qquad $V^{p}=V^{p}\cup\{{v_{m}^{p}}\}$;\\
    \qquad\qquad $u_{mw}^{p}=1, \text{\,and\,}u_{mj}^{p}=0, \text{\,with\,}j=1,2...n,j\neq{w}$ \\
    \qquad \textbf{Until}(m=k) \\
\textbf{end for}\\
\qquad $U^{\ast}$ = $\frac{1}{P}\sum\limits_{p=1}^{P}U^{p}$\\
\hline
\end{tabular}
\end{center}
\subsection{Results on data sets}
We partition each data set randomly into equal sized chunks. The size of each chunk can be specified by the user. Normally it refers to a certain percentage of the entire data size. The size of the last chunk maybe smaller than the others if the entire data set can not be divided by the chunk size. We conduct experiments with chunk sizes as 1\%, 2.5\%, 5\%, 10\% and 25\% of the entire data set size for MF, IS, Caltech7, Caltech20 and Reuters data sets. For Forest data set, limited by memory, smaller percentages are chosen for chunk sizes as 0.1\%, 0.25\%, 0.5\%, 1\% and 2.5\% of the entire data set size. For each data set, the same fuzzifier m is set for all the approaches and 20 trials are run with random order of the input data. We calculate the mean, standard deviation of the values of accuracy, NMI and F-measure over 20 trials.
The results of the six data sets are shown in Table.~\ref{table:MF}, Table.~\ref{table:IS}, Table.~\ref{table:Caltech7}, Table.~\ref{table:Caltech20}, Table.~\ref{table:Reuters} and Table.~\ref{table:Forest} respectively. From the tables we can see, IminimaxFCM1 and IminimaxFCM2 always produce the best partition every time with various chunk sizes.

Table.~\ref{table:MF} (a), (b), (c) are the accuracy, NMI and F-measure results on MF data set respectively. The results show that our multi-view based proposed approaches IminimaxFCM1 and  IminimaxFCM2 perform better than single concatenated view based approaches OFCM and SPFCM. Compared with OFCM, the improvements of IminimaxFCM1 on average of accuracy, NMI and F-measure over all the chunk sizes are $11.9\%$, $15.8\%$ and $11.3\%$ and the improvements of IminimaxFCM2 are $15.8\%$, $17.2\%$ and $13.8\%$. Compared with SPFCM, the improvements of IminimaxFCM1 and IminimaxFCM2 are $23.2\%$, $24.3\%$, $21.6\%$ and $27.5\%$, $25.9\%$, $24.4\%$ respectively. Moreover, the results show that IminimaxFCM1 and IminimaxFCM2 by using minimax optimization to integrate different views also perform better than NaiveMVOFCM and NaiveMVSPFCM in which each view is processed individually without considering the complementary information among different views. Compared with NaiveMVOFCM, the improvements of IminimaxFCM1 and IminimaxFCM2 are $44.8\%$, $47.6\%$, $42.0\%$ and $49.8\%$, $49.5\%$, $45.2\%$ respectively. Compare with NaiveMVSPFCM, the improvements of IminimaxFCM1 and IminimaxFCM2 are $49.4\%$, $51.7\%$, $46.3\%$ and $54.6\%$, $53.6\%$, $49.6\%$ respectively. The results on the other data sets provide similar pattern in performance and the best performance always achieved by IminimaxFCM1 or IminimaxFCM2.

\begin{sidewaystable}
\scriptsize
\caption{Comparison of related incremental clustering approaches on MF} 
\centering 
\subfloat[][Accuracy]{
\begin{tabular}{c c c c c c c}
\hline
 Chunk Size& OFCM & SPFCM & NaiveMVOFCM & NaiveMVSPFCM & IMinimaxFCM1 & IMinimaxFCM2 \\ [0.5ex] 
\hline 
$1\%$ & 0.8196(0.0538)  & $0.6922(0.0491)$ & $0.6264(0.0665)$ & $0.5888(0.0597)$ & $0.8942(0.0676)$ & $\textbf{0.9581}(0.0014)$\\
$2.5\%$ & 0.8149(0.0796)  & $0.7160(0.0668)$ & $0.6184(0.0715)$ & $0.5990(0.0656)$ & $0.9101(0.0644)$ & $\textbf{0.9479}(0.0379)$\\
$5\%$ & 0.8162(0.0672)  & $0.7347(0.0704)$ & $0.6594(0.0661)$ & $0.6034(0.0379)$ & $0.8965(0.0643)$ & $\textbf{0.9271}(0.0517)$\\
$10\%$ & 0.8143(0.0678)  & $0.7688(0.0733)$ & $0.6092(0.0447)$ & $0.6238(0.0561)$ & $0.9375(0.0482)$ & $\textbf{0.9477}(0.0413)$\\
$25\%$ & 0.8081(0.0693)  & $0.7868(0.0919)$ & $0.6342(0.0598)$ & $0.6358(0.0641)$ & $0.9190(0.0628)$ & $\textbf{0.9348}(0.0522)$\\
\hline 
\end{tabular}}

\subfloat[][NMI]{
\begin{tabular}{c c c c c c c}
\hline
 Chunk Size& OFCM & SPFCM & NaiveMVOFCM & NaiveMVSPFCM & IMinimaxFCM1 & IMinimaxFCM2 \\ [0.5ex] 
\hline 
$1\%$ & 0.7674(0.0324)  & $0.6830(0.0181)$ & $0.5989(0.0277)$ & $0.5699(0.0395)$ & $0.8888(0.0218)$ & $\textbf{0.9104}(0.0019)$\\
$2.5\%$ & 0.7722(0.0388)  & $0.7016(0.0370)$ & $0.6097(0.0439)$ & $0.5793(0.0316)$ & $0.8956(0.0220)$ & $\textbf{0.9100}(0.0123)$\\
$5\%$ & 0.7745(0.0330)  & $0.7154(0.0468)$ & $0.6198(0.0257)$ & $0.5837(0.0293)$ & $0.8902(0.0194)$ & $\textbf{0.9004}(0.0204)$\\
$10\%$ & 0.7727(0.0321)  & $0.7484(0.0361)$ & $0.5908(0.0326)$ & $0.5990(0.0419)$ & $0.9050(0.0173)$ & $\textbf{0.9102}(0.0161)$\\
$25\%$ & 0.7819(0.0342)  & $0.7536(0.0512)$ & $0.6145(0.0414)$ & $0.6210(0.0402)$ & $0.8985(0.0169)$ & $\textbf{0.9039}(0.0178)$\\
\hline 
\end{tabular}}

\subfloat[][F-measure]{
\begin{tabular}{c c c c c c c}
\hline
 Chunk Size& OFCM & SPFCM & NaiveMVOFCM & NaiveMVSPFCM & IMinimaxFCM1 & IMinimaxFCM2 \\ [0.5ex] 
\hline 
$1\%$ & 0.8316(0.0423)  & $0.7248(0.0288)$ & $0.6456(0.0569)$ & $0.6131(0.0525)$ & $0.9155(0.0451)$ & $\textbf{0.9580}(0.0015)$\\
$2.5\%$ & 0.8364(0.0578)  & $0.7440(0.0521)$ & $0.6534(0.0643)$ & $0.6214(0.0497)$ & $0.9278(0.0461)$ & $\textbf{0.9523}(0.0241)$\\
$5\%$ & 0.8358(0.0483)  & $0.7517(0.0650)$ & $0.6753(0.0424)$ & $0.6248(0.0349)$ & $0.9186(0.0413)$ & $\textbf{0.9386}(0.0336)$\\
$10\%$ & 0.8323(0.0478)  & $0.7913(0.0524)$ & $0.6342(0.0361)$ & $0.6465(0.0523)$ & $0.9451(0.0319)$ & $\textbf{0.9523}(0.0271)$\\
$25\%$ & 0.8322(0.0512)  & $0.8041(0.0748)$ & $0.6593(0.0539)$ & $0.6667(0.0595)$ & $0.9336(0.0393)$ & $\textbf{0.9439}(0.0336)$\\
\hline 
\end{tabular}}
\label{table:MF} 
\end{sidewaystable}

\begin{sidewaystable}[!htbp]
\scriptsize
\caption{Comparison of related incremental clustering approaches on IS} 
\centering 
\subfloat[][Accuracy]{
\begin{tabular}{c c c c c c c}
\hline
 Chunk Size& OFCM & SPFCM & NaiveMVOFCM & NaiveMVSPFCM & IMinimaxFCM1 & IMinimaxFCM2 \\ [0.5ex] 
\hline 
$1\%$ & 0.6001(0.0257)  & $0.5795(0.0581)$ & $0.5670(0.0599)$ & $0.5596(0.0559)$ & $0.6100(0.0198)$ & $\textbf{0.6192}(0.0143)$\\
$2.5\%$ & 0.6275(0.0246)  & $0.6068(0.0423)$ & $0.5098(0.0666)$ & $0.5552(0.0798)$ & $0.6411(0.0298)$ & $\textbf{0.6441}(0.0403)$\\
$5\%$ & 0.6177(0.0373)  & $0.6343(0.0314)$ & $0.5219(0.0695)$ & $0.5217(0.0603)$ & $0.6692(0.0452)$ & $\textbf{0.6774}(0.0387)$\\
$10\%$ & 0.6091(0.0350)  & $0.6117(0.0376)$ & $0.5285(0.0432)$ & $0.5497(0.0953)$ & $\textbf{0.6861}(0.0244)$ & $0.6568(0.0518)$\\
$25\%$ & 0.5957(0.0489)  & $0.6392(0.0459)$ & $0.5019(0.0670)$ & $0.5568(0.0338)$ & $0.6900(0.0236)$ & $\textbf{0.6992}(0.0343)$\\
\hline 
\end{tabular}}

\subfloat[][NMI]{
\begin{tabular}{c c c c c c c}
\hline
 Chunk Size& OFCM & SPFCM & NaiveMVOFCM & NaiveMVSPFCM & IMinimaxFCM1 & IMinimaxFCM2 \\ [0.5ex] 
\hline 
$1\%$ & 0.5786(0.0151)  & $0.5842(0.0358)$ & $0.5459(0.0789)$ & $0.5371(0.0482)$ & $0.5917(0.0120)$ & $\textbf{0.5922}(0.0081)$\\
$2.5\%$ & 0.5964(0.0290)  & $0.5916(0.0306)$ & $0.4807(0.0460)$ & $0.5390(0.0714)$ & $0.6089(0.0165)$ & $\textbf{0.6096}(0.0238)$\\
$5\%$ & 0.6163(0.0326)  & $0.6128(0.0299)$ & $0.4715(0.0756)$ & $0.5178(0.0683)$ & $0.6131(0.0229)$ & $\textbf{0.6173}(0.0146)$\\
$10\%$ & 0.5955(0.0463)  & $0.6023(0.0281)$ & $0.5064(0.0590)$ & $0.5236(0.0794)$ & $\textbf{0.6309}(0.0039)$ & $0.6108(0.0162)$\\
$25\%$ & 0.5755(0.0637)  & $0.6119(0.0344)$ & $0.4675(0.0734)$ & $0.5239(0.0474)$ & $0.6146(0.0126)$ & $\textbf{0.6186}(0.0105)$\\
\hline 
\end{tabular}}

\subfloat[][F-measure]{
\begin{tabular}{c c c c c c c}
\hline
 Chunk Size& OFCM & SPFCM & NaiveMVOFCM & NaiveMVSPFCM & IMinimaxFCM1 & IMinimaxFCM2 \\ [0.5ex] 
\hline 
$1\%$ & 0.6179(0.0149)  & $0.6192(0.0450)$ & $0.6120(0.0635)$ & $0.5993(0.0578)$ & $0.6236(0.0152)$ & $\textbf{0.6304}(0.0087)$\\
$2.5\%$ & 0.6568(0.0181)  & $0.6353(0.0324)$ & $0.5530(0.0550)$ & $0.6056(0.0690)$ & $0.6425(0.0275)$ & $\textbf{0.6588}(0.0389)$\\
$5\%$ & 0.6654(0.0256)  & $0.6637(0.0380)$ & $0.5581(0.0626)$ & $0.5809(0.0507)$ & $0.6661(0.0454)$ & $\textbf{0.6842}(0.0278)$\\
$10\%$ & 0.6542(0.0389)  & $0.6539(0.0334)$ & $0.5820(0.0426)$ & $0.5920(0.0797)$ & $\textbf{0.6961}(0.0085)$ & $0.6706(0.0353)$\\
$25\%$ & 0.6354(0.0600)  & $0.6622(0.0422)$ & $0.5403(0.0672)$ & $0.5949(0.0394)$ & $0.6840(0.0280)$ & $\textbf{0.7000}(0.0247)$\\
\hline 
\end{tabular}}

\label{table:IS} 
\end{sidewaystable}

\begin{sidewaystable}[!htbp]
\scriptsize
\caption{Comparison of related incremental clustering approaches on Caltech7} 
\centering 
\subfloat[][Accuracy]{
\begin{tabular}{c c c c c c c}
\hline
 Chunk Size& OFCM & SPFCM & NaiveMVOFCM & NaiveMVSPFCM & IMinimaxFCM1 & IMinimaxFCM2 \\ [0.5ex] 
\hline 
$1\%$ & 0.5104(0.0592)  & $0.5355(0.1012)$ & $0.5714(0.1335)$ & $0.5323(0.0553)$ & $\textbf{0.6414}(0.0267)$ & $0.5695(0.0689)$\\
$2.5\%$ & 0.5851(0.0089)  & $0.5227(0.0749)$ & $0.5513(0.1389)$ & $0.5507(0.1168)$ & $\textbf{0.6483}(0.0355)$ & $0.5862(0.0915)$\\
$5\%$ & 0.5331(0.0523)  & $0.5289(0.0905)$ & $0.5347(0.0552)$ & $0.5182(0.1024)$ & $\textbf{0.7000}(0.0536)$ & $0.6524(0.0748)$\\
$10\%$ & 0.4802(0.0556)  & $0.5388(0.1188)$ & $0.5877(0.0615)$ & $0.5704(0.1193)$ & $\textbf{0.6362}(0.0415)$ & $0.5942(0.0785)$\\
$25\%$ & 0.4524(0.0554)  & $0.5132(0.0491)$ & $0.5763(0.1255)$ & $0.5326(0.1193)$ & $0.5970(0.0192)$ & $\textbf{0.6132}(0.0766)$\\
\hline 
\end{tabular}}

\subfloat[][NMI]{
\begin{tabular}{c c c c c c c}
\hline
 Chunk Size& OFCM & SPFCM & NaiveMVOFCM & NaiveMVSPFCM & IMinimaxFCM1 & IMinimaxFCM2 \\ [0.5ex] 
\hline 
$1\%$ & 0.5629(0.0106)  & $0.5020(0.0438)$ & $0.4535(0.0756)$ & $0.3354(0.0514)$ & $\textbf{0.5700}(0.0112)$ & $0.5632(0.0409)$\\
$2.5\%$ & 0.5719(0.0303)  & $0.5196(0.0190)$ & $0.3502(0.0922)$ & $0.3590(0.0862)$ & $\textbf{0.5955}(0.0293)$ & $0.5631(0.0397)$\\
$5\%$ & 0.5595(0.0175) & $0.5372(0.0276)$ & $0.3539(0.0880)$ & $0.3817(0.0875)$ & $\textbf{0.5588}(0.0477)$ & $0.5505(0.0345)$\\
$10\%$ & 0.5471(0.0278)  & $0.5245(0.0258)$ & $0.3901(0.0272)$ & $0.3625(0.0907)$ & $0.5457(0.0462)$ & $\textbf{0.5800}(0.0285)$\\
$25\%$ & 0.5521(0.0203)  & $0.5479(0.0247)$ & $0.3550(0.0179)$ & $0.3824(0.0865)$ & $0.5775(0.0217)$ & $\textbf{0.5953}(0.0239)$\\
\hline 
\end{tabular}}

\subfloat[][F-measure]{
\begin{tabular}{c c c c c c c}
\hline
 Chunk Size& OFCM & SPFCM & NaiveMVOFCM & NaiveMVSPFCM & IMinimaxFCM1 & IMinimaxFCM2 \\ [0.5ex] 
\hline 
$1\%$ & 0.6291(0.0451)  & $0.6303(0.0892)$ & $0.6605(0.1071)$ & $0.5998(0.0504)$ & $\textbf{0.7384}(0.0198)$ & $0.6798(0.0560)$\\
$2.5\%$ & 0.6785(0.0112)  & $0.6326(0.0543)$ & $0.6062(0.1173)$ & $0.6231(0.0885)$ & $\textbf{0.7462}(0.0309)$ & $0.6905(0.0690)$\\
$5\%$ & 0.6399(0.0407)  & $0.6527(0.0658)$ & $0.6022(0.0698)$ & $0.5933(0.1031)$ & $\textbf{0.7704}(0.0408)$ & $0.7387(0.0543)$\\
$10\%$ & 0.6046(0.0378)  & $0.6414(0.0895)$ & $0.6477(0.0432)$ & $0.6379(0.1051)$ & $\textbf{0.7242}(0.0409)$ & $0.6940(0.0652)$\\
$25\%$ & 0.5872(0.0411)  & $0.6231(0.0294)$ & $0.6368(0.0866)$ & $0.6140(0.0953)$ & $0.6874(0.0149)$ & $\textbf{0.7081}(0.0605)$\\
\hline 
\end{tabular}}

\label{table:Caltech7} 
\end{sidewaystable}

\begin{sidewaystable}[!htbp]
\scriptsize
\caption{Comparison of related incremental clustering approaches on Caltech20} 
\centering 
\subfloat[][Accuracy]{
\begin{tabular}{c c c c c c c}
\hline
 Chunk Size& OFCM & SPFCM & NaiveMVOFCM & NaiveMVSPFCM & IMinimaxFCM1 & IMinimaxFCM2 \\ [0.5ex] 
\hline 
$1\%$ & 0.4995(0.0510)  & $0.4511(0.0552)$ & $0.4212(0.0814)$ & $0.4039(0.0845)$ & $\textbf{0.5516}(0.0585)$ & $0.5109(0.0346)$\\
$2.5\%$ & 0.5181(0.0235)  & $0.4671(0.0340)$ & $0.4233(0.0650)$ & $0.3711(0.0611)$ & $\textbf{0.5830}(0.0683)$ & $0.5253(0.0170)$\\
$5\%$ & 0.5299(0.0383)  & $0.4577(0.0475)$ & $0.4220(0.0560)$ & $0.4095(0.0868)$ & $\textbf{0.5596}(0.0381)$ & $0.5396(0.0222)$\\
$10\%$ & 0.4925(0.0243)  & $0.4640(0.0516)$ & $0.4030(0.0511)$ & $0.3918(0.0937)$ & $\textbf{0.5788}(0.0477)$ & $0.5438(0.0297)$\\
$25\%$ & 0.4640(0.0456)  & $0.4410(0.0475)$ & $0.4277(0.0705)$ & $0.3930(0.0917)$ & $\textbf{0.5781}(0.0619)$ & $0.5513(0.0667)$\\
\hline 
\end{tabular}}

\subfloat[][NMI]{
\begin{tabular}{c c c c c c c}
\hline
 Chunk Size& OFCM & SPFCM & NaiveMVOFCM & NaiveMVSPFCM & IMinimaxFCM1 & IMinimaxFCM2 \\ [0.5ex] 
\hline 
$1\%$ & 0.6139(0.0177)  & $0.5744(0.0217)$ & $0.3723(0.0498)$ & $0.3477(0.0586)$ & $\textbf{0.6487}(0.0169)$ & $0.6280(0.0218)$\\
$2.5\%$ & 0.6009(0.0053)  & $0.5711(0.0147)$ & $0.3755(0.0346)$ & $0.3604(0.0483)$ & $\textbf{0.6359}(0.0199)$ & $0.6309(0.0078)$\\
$5\%$ & 0.6047(0.0060) & $0.5820(0.0183)$ & $0.3695(0.0438)$ & $0.3729(0.0397)$ & $\textbf{0.6114}(0.0159)$ & $0.6166(0.0197)$\\
$10\%$ & 0.6150(0.0090)  & $0.5739(0.0227)$ & $0.3831(0.0383)$ & $0.3706(0.0527)$ & $\textbf{0.6189}(0.0074)$ & $0.6117(0.0127)$\\
$25\%$ & 0.6146(0.0185)  & $0.5768(0.0252)$ & $0.3918(0.0236)$ & $0.3745(0.0447)$ & $\textbf{0.6300}(0.0196)$ & $0.6227(0.0232)$\\
\hline 
\end{tabular}}

\subfloat[][F-measure]{
\begin{tabular}{c c c c c c c}
\hline
 Chunk Size& OFCM & SPFCM & NaiveMVOFCM & NaiveMVSPFCM & IMinimaxFCM1 & IMinimaxFCM2 \\ [0.5ex] 
\hline 
$1\%$ & 0.5862(0.0428)  & $0.5344(0.0527)$ & $0.4458(0.0642)$ & $0.4354(0.0776)$ & $\textbf{0.6363}(0.0498)$ & $0.5958(0.0340)$\\
$2.5\%$ & 0.5898(0.0200)  & $0.5484(0.0281)$ & $0.4578(0.0712)$ & $0.4093(0.0553)$ & $\textbf{0.6560}(0.0555)$ & $0.6104(0.0190)$\\
$5\%$ & 0.6025(0.0321)  & $0.5439(0.0464)$ & $0.4417(0.0494)$ & $0.4415(0.0735)$ & $\textbf{0.6293}(0.0352)$ & $0.6097(0.0202)$\\
$10\%$ & 0.5780(0.0222)  & $0.5431(0.0479)$ & $0.4486(0.0479)$ & $0.4301(0.0830)$ & $\textbf{0.6399}(0.0333)$ & $0.6170(0.0363)$\\
$25\%$ & 0.5621(0.0417)  & $0.5265(0.0468)$ & $0.4709(0.0559)$ & $0.4338(0.0827)$ & $\textbf{0.6466}(0.0511)$ & $0.6212(0.0562)$\\
\hline 
\end{tabular}}

\label{table:Caltech20} 
\end{sidewaystable}

\begin{sidewaystable}[!htbp]
\scriptsize
\caption{Comparison of related incremental clustering approaches on Reuters} 
\centering 
\subfloat[][Accuracy]{
\begin{tabular}{c c c c c c c}
\hline
 Chunk Size& OFCM & SPFCM & NaiveMVOFCM & NaiveMVSPFCM & IMinimaxFCM1 & IMinimaxFCM2 \\ [0.5ex] 
\hline 
$1\%$ & 0.4700(0.0644)  & $0.4371(0.0448)$ & $0.3572(0.0455)$ & $0.3191(0.0188)$ & $\textbf{0.5009}(0.0285)$ & $0.4936(0.0135)$\\
$2.5\%$ & 0.4741(0.0668)  & $0.4525(0.0374)$ & $0.3430(0.0488)$ & $0.3325(0.0356)$ & $\textbf{0.5012}(0.0197)$ & $0.4993(0.0148)$\\
$5\%$ & 0.4979(0.0085)  & $0.4455(0.0543)$ & $0.3457(0.0322)$ & $0.3429(0.0394)$ & $\textbf{0.4987}(0.0113)$ & $0.4921(0.0028)$\\
$10\%$ & 0.4767(0.0422)  & $0.4452(0.0453)$ & $0.3250(0.0268)$ & $0.2981(0.0219)$ & $0.4983(0.0135)$ & $\textbf{0.5028}(0.0090)$\\
$25\%$ & 0.4851(0.0262)  & $0.4458(0.0341)$ & $0.3223(0.0407)$ & $0.3182(0.0383)$ & $\textbf{0.4923}(0.0105)$ & $0.4980(0.0181)$\\
\hline 
\end{tabular}}

\subfloat[][NMI]{
\begin{tabular}{c c c c c c c}
\hline
 Chunk Size& OFCM & SPFCM & NaiveMVOFCM & NaiveMVSPFCM & IMinimaxFCM1 & IMinimaxFCM2 \\ [0.5ex] 
\hline 
$1\%$ & 0.2731(0.0341)  & $0.2303(0.0320)$ & $0.1298(0.0343)$ & $0.0847(0.0203)$ & $\textbf{0.2900}(0.0073)$ & $0.2898(0.0075)$\\
$2.5\%$ & 0.2626(0.0345)  & $0.2461(0.0267)$ & $0.1235(0.0392)$ & $0.0974(0.0293)$ & $\textbf{0.2938}(0.0037)$ & $0.2897(0.0094)$\\
$5\%$ & 0.2897(0.0122) & $0.2414(0.0389)$ & $0.1088(0.0219)$ & $0.1048(0.0219)$ & $\textbf{0.2940}(0.0102)$ & $0.2945(0.0030)$\\
$10\%$ & 0.2646(0.0243)  & $0.2523(0.0288)$ & $0.0958(0.0175)$ & $0.0784(0.0106)$ & $0.2988(0.0052)$ & $\textbf{0.2998}(0.0073)$\\
$25\%$ & 0.2880(0.0134)  & $0.2467(0.0209)$ & $0.0912(0.0334)$ & $0.0907(0.0291)$ & $\textbf{0.2978}(0.0062)$ & $0.2892(0.0085)$\\
\hline 
\end{tabular}}

\subfloat[][F-measure]{
\begin{tabular}{c c c c c c c}
\hline
 Chunk Size& OFCM & SPFCM & NaiveMVOFCM & NaiveMVSPFCM & IMinimaxFCM1 & IMinimaxFCM2 \\ [0.5ex] 
\hline 
$1\%$ & 0.5023(0.0453)  & $0.4596(0.0396)$ & $0.3681(0.0442)$ & $0.3314(0.0253)$ & $\textbf{0.5269}(0.0129)$ & $0.5213(0.0093)$\\
$2.5\%$ & 0.5029(0.0435)  & $0.4746(0.0345)$ & $0.3538(0.0417)$ & $0.3369(0.0340)$ & $\textbf{0.5250}(0.0065)$ & $0.5228(0.0119)$\\
$5\%$ & 0.5222(0.0098)  & $0.4700(0.0484)$ & $0.3475(0.0288)$ & $0.3468(0.0294)$ & $\textbf{0.5261}(0.0106)$ & $0.5199(0.0045)$\\
$10\%$ & 0.4974(0.0328)  & $0.4765(0.0356)$ & $0.3334(0.0227)$ & $0.3097(0.0238)$ & $0.5267(0.0057)$ & $0.\textbf{5274}(0.0063)$\\
$25\%$ & 0.5141(0.0254)  & $0.4734(0.0295)$ & $0.3260(0.0371)$ & $0.3255(0.0399)$ & $\textbf{0.5193}(0.0051)$ & $0.5161(0.0128)$\\
\hline 
\end{tabular}}

\label{table:Reuters} 
\end{sidewaystable}

\begin{sidewaystable}[!htbp]
\scriptsize
\caption{Comparison of related incremental clustering approaches on Forest} 
\centering 
\subfloat[][Accuracy]{
\begin{tabular}{c c c c c c c}
\hline
 Chunk Size& OFCM & SPFCM & NaiveMVOFCM & NaiveMVSPFCM & IMinimaxFCM1 & IMinimaxFCM2 \\ [0.5ex] 
\hline 
$1\%$ & 0.3428(0.0219)  & $0.3234(0.0305)$ & $0.3211(0.0546)$ & $0.3318(0.0503)$ & $\textbf{0.4401}(0.0060)$ & $0.3592(0.0347)$\\
$2.5\%$ & 0.3368(0.0203)  & $0.3123(0.0281)$ & $0.3249(0.0278)$ & $0.3443(0.0267)$ & $\textbf{0.4113}(0.0337)$ & $0.3458(0.0423)$\\
$5\%$ & 0.3417(0.0187)  & $0.3037(0.0270)$ & $0.3441(0.0266)$ & $0.3020(0.0225)$ & $\textbf{0.4292}(0.0288)$ & $0.3846(0.0003)$\\
$10\%$ & 0.3464(0.0155)  & $0.3154(0.0313)$ & $0.3187(0.0580)$ & $0.3189(0.0275)$ & $\textbf{0.4148}(0.0484)$ & $0.3899(0.0676)$\\
$25\%$ & 0.3438(0.0165)  & $0.3106(0.0257)$ & $0.3591(0.0570)$ & $0.3494(0.0507)$ & $\textbf{0.3866}(0.0566)$ & $0.3595(0.0360)$\\
\hline 
\end{tabular}}

\subfloat[][NMI]{
\begin{tabular}{c c c c c c c}
\hline
 Chunk Size& OFCM & SPFCM & NaiveMVOFCM & NaiveMVSPFCM & IMinimaxFCM1 & IMinimaxFCM2 \\ [0.5ex] 
\hline 
$1\%$ & 0.0927(0.0118)  & $0.1067(0.0091)$ & $0.1079(0.0259)$ & $0.1156(0.0156)$ & $\textbf{0.1669}(0.0024)$ & $0.1462(0.0054)$\\
$2.5\%$ & 0.0774(0.0180)  & $0.1128(0.0147)$ & $0.1082(0.0154)$ & $0.0931(0.0192)$ & $\textbf{0.1592}(0.0101)$ & $0.1442(0.0069)$\\
$5\%$ & 0.0799(0.0148) & $0.1070(0.0101)$ & $0.1151(0.0151)$ & $0.1035(0.0203)$ & $\textbf{0.1650}(0.0086)$ & $0.1506(0.0002)$\\
$10\%$ & 0.0818(0.0157)  & $0.1033(0.0096)$ & $0.0969(0.0334)$ & $0.0972(0.0208)$ & $0.1238(0.0679)$ & $\textbf{0.1565}(0.0168)$\\
$25\%$ & 0.0845(0.0145)  & $0.1059(0.0105)$ & $0.1156(0.0207)$ & $0.1199(0.0196)$ & $\textbf{0.1535}(0.0143)$ & $0.1425(0.0056)$\\
\hline 
\end{tabular}}

\subfloat[][F-measure]{
\begin{tabular}{c c c c c c c}
\hline
 Chunk Size& OFCM & SPFCM & NaiveMVOFCM & NaiveMVSPFCM & IMinimaxFCM1 & IMinimaxFCM2 \\ [0.5ex] 
\hline 
$1\%$ & 0.3942(0.0105)  & $0.3732(0.0259)$ & $0.3838(0.0494)$ & $0.3858(0.0391)$ & $\textbf{0.4772}(0.0009)$ & $0.4614(0.0357)$\\
$2.5\%$ & 0.3371(0.0123)  & $0.3682(0.0202)$ & $0.4057(0.0193)$ & $0.3857(0.0145)$ & $\textbf{0.4656}(0.0276)$ & $0.4346(0.0402)$\\
$5\%$ & 0.3915(0.0116)  & $0.3587(0.0233)$ & $0.4106(0.0198)$ & $0.3651(0.0215)$ & $\textbf{0.4770}(0.0002)$ & $0.4165(0.0004)$\\
$10\%$ & 0.3921(0.0058)  & $0.3659(0.0254)$ & $0.3714(0.0485)$ & $0.3812(0.0466)$ & $\textbf{0.4626}(0.0464)$ & $0.4581(0.0374)$\\
$25\%$ & 0.3891(0.0149)  & $0.3646(0.0222)$ & $0.4172(0.0346)$ & $0.4166(0.0447)$ & $\textbf{0.4608}(0.0320)$ & $0.4290(0.0412)$\\
\hline 
\end{tabular}}

\label{table:Forest} 
\end{sidewaystable}

\section{Conclusion}
We have proposed a new incremental fuzzy clustering approach called IminimaxFCM for large multi-view data analysis, and apply IminimaxFCM on six multi-view data sets to demonstrate its feasibility and potential. IminimaxFCM processes large multi-view data chunk by chunk. One distinctive characteristic of IminimaxFCM is that in each chunk minimax optimization is applied to integrate different views and the multi-view centroids are identified to represent that chunk. Experiments conducted on six real world multi-view data sets including image and document data sets show that IminimaxFCM not only outperforms related incremental single view based algorithms with more accurate clustering results, but also performs better than two incremental multi-view based algorithms in which each view is processed individually. The merits showed in the experiments indicate that IminimaxFCM has a great potential to be used for large multi-view data clustering. In the future, much larger data sets need to be collected and tested. Further experimental studies may be conducted on real world big data by implementing the approach on the Hadoop/Spark framework.

\section*{References}
\bibliography{reference}

\end{document}